\newcommand{\PaperTitle}{SHELF: A Synthetic Harness for Multi-Task Bibliographic Benchmarking}

\newcommand{\PaperShortTitle}{SHELF}
\newcommand{\PaperDate}{\today}
\newcommand{\PaperAbstract}{Libraries and archives manage large collections with limited staff and computing budgets, yet common benchmarks do not systematically test their bibliographic work. They need to know which methods work for their tasks and what those methods require to run. SHELF, the Synthetic Harness for Evaluating LLM Fitness, addresses this gap. It is a Python system that turns labelled taxonomies, writing specifications, and a generation budget into controlled benchmark data and evaluation tasks. This first release contains 62,899 model-written documents based on Library of Congress vocabularies, with tasks for classification, clustering, retrieval, pair classification, and instruction retrieval. We compare TF, TF-IDF, BM25, popular encoders, and, on subject classification only, zero-shot decoders; each method appears only on tasks that support it. Subject classification reaches 0.8887, while genre-form classification reaches only 0.2605, and several pair and clustering tasks remain near chance. Sparse methods remain competitive on classification, while TF-IDF is the fastest measured arm in the subject timing experiment. SHELF also varies bibliographic facets independently and can generate new, verifiably unseen documents after a model's training cutoff. Comparisons with LCSHBench and Project Gutenberg show that model rankings transfer more reliably than absolute scores, but SHELF scores do not estimate accuracy on production catalogue data. We release all source code and data under permissive licenses on GitHub and Hugging Face.}
\newcommand{\PaperKeywords}{bibliographic classification, text embeddings, benchmark validity, synthetic evaluation data, Library of Congress Classification, transfer}

\newcommand{\PaperFontProfile}{libertinus}
\newcommand{\PaperLineSpacing}{single}

\newcommand{\PaperBibTitle}{References}

\newcommand{\PaperAIStatement}{Portions of this work were prepared with assistance from large language models. The author is solely responsible for all content, including any errors or omissions.}
\newcommand{\PaperFunding}{none to declare}
\newcommand{\PaperCompeting}{none to declare}

\newcommand{\PaperBaseSize}{10pt}
\newcommand{\PaperPaperOption}{letterpaper}

\newcommand{\PaperAuthorPlain}{Michael J. Bommarito II}

\newif\ifPaperHasSubtitle\PaperHasSubtitlefalse
\newif\ifPaperHasKeywords\PaperHasKeywordstrue
\newif\ifPaperHasJEL\PaperHasJELtrue
\newif\ifPaperHasMSC\PaperHasMSCfalse
\newif\ifPaperHasSeries\PaperHasSeriesfalse
\newif\ifPaperVenueArxiv\PaperVenueArxivtrue
\newif\ifPaperVenueSSRN\PaperVenueSSRNfalse
\newif\ifPaperVenuePreprint\PaperVenuePreprintfalse
\newif\ifPaperEnginePDF\PaperEnginePDFtrue
\newif\ifPaperBibNatbib\PaperBibNatbibtrue
\newif\ifPaperBibBiblatex\PaperBibBiblatexfalse
\newif\ifPaperBibNumeric\PaperBibNumerictrue
\newif\ifPaperTwoColumn\PaperTwoColumntrue
\newif\ifPaperHasTitleBackground\PaperHasTitleBackgroundfalse
\newif\ifPaperModuleBoxes\PaperModuleBoxestrue
\newif\ifPaperModuleCode\PaperModuleCodetrue
\newif\ifPaperModuleAlgo\PaperModuleAlgotrue
\newif\ifPaperModuleSI\PaperModuleSItrue
\newif\ifPaperSortCites\PaperSortCitestrue
\newif\ifPaperAIUsed\PaperAIUsedtrue
\newcommand{\PaperAuthorBlock}{%
  {\large Michael J. Bommarito II\footnotemark\par}%
  \vspace{0.35em}{\small\texttt{michael.bommarito@gmail.com}\par}%
}
\newcommand{\PaperAuthorFootnote}{%
  \footnotetext{Portions of this work were prepared with assistance from large language models. The author is solely responsible for all content, including any errors or omissions.}%
}

  \documentclass[\PaperBaseSize,\PaperPaperOption,twocolumn]{article}

\usepackage{etoolbox}
\usepackage{iftex}

  \usepackage[T1]{fontenc}
  \usepackage[utf8]{inputenc}

\usepackage[american]{babel}
\usepackage{csquotes}   %

\usepackage[\PaperPaperOption,margin=1in]{geometry}
\usepackage{setspace}       %
\usepackage{fancyhdr}       %
\usepackage{titlesec}       %
\usepackage{titling}        %
\usepackage[bottom]{footmisc} %

\usepackage{amsmath,amssymb}
\usepackage{mathtools}
\usepackage{amsthm}

\usepackage{graphicx}
\graphicspath{{figures/}}          %
\usepackage{xcolor}                %
\usepackage{float}
\usepackage{rotating}              %
\usepackage{pdflscape}             %
\usepackage[labelfont=bf,font=small,skip=6pt]{caption}
\usepackage{subcaption}

\usepackage{booktabs}              %
\usepackage{tabularx}
\usepackage{longtable}             %
\usepackage{threeparttable}        %
\usepackage{multirow}
\usepackage{makecell}

\usepackage{eso-pic}

\usepackage{enumitem}

  \usepackage{siunitx}
      \usepackage[numbers,sort&compress,square,comma]{natbib}

  \newcommand{\PrintBibliography}{%
    \phantomsection\addcontentsline{toc}{section}{\PaperBibTitle}%
    \renewcommand{\refname}{\PaperBibTitle}%
    \bibliography{bib/references}}

  \ifdefstring{\PaperFontProfile}{libertinus}{%
    \usepackage{libertinus}%
    \usepackage[varqu,scaled=0.96]{zi4}%
  }{}%
  \ifdefstring{\PaperFontProfile}{newtx}{%
    \usepackage{newtxtext}%
    \usepackage{newtxmath}%
    \usepackage[varqu,scaled=0.95]{zi4}%
  }{}%
  \ifdefstring{\PaperFontProfile}{lmodern}{%
    \usepackage{lmodern}%
  }{}%
  \ifdefstring{\PaperFontProfile}{plex}{%
    \PackageError{paper-template}%
      {font_profile 'plex' is OpenType-only}%
      {Set typography.engine to xelatex or lualatex, or pick another profile.}%
  }{}%

  \usepackage[protrusion=true,expansion=true,factor=1100,final]{microtype}

\definecolor{gray-50}{HTML}{FAFAFA}
\definecolor{gray-100}{HTML}{F5F5F5}
\definecolor{gray-200}{HTML}{E5E5E5}
\definecolor{gray-300}{HTML}{D4D4D4}
\definecolor{gray-400}{HTML}{A3A3A3}
\definecolor{gray-500}{HTML}{737373}
\definecolor{gray-600}{HTML}{525252}
\definecolor{gray-700}{HTML}{404040}
\definecolor{gray-800}{HTML}{262626}
\definecolor{gray-900}{HTML}{171717}
\definecolor{gray-950}{HTML}{0A0A0A}

\definecolor{accent-50}{HTML}{EFF3F8}
\definecolor{accent-100}{HTML}{D6E0EC}
\definecolor{accent-200}{HTML}{ADC1D9}
\definecolor{accent-300}{HTML}{7C9BBF}
\definecolor{accent-400}{HTML}{4E73A0}
\definecolor{accent-500}{HTML}{2F587F}   %
\definecolor{accent-600}{HTML}{264964}
\definecolor{accent-700}{HTML}{1E3A5F}
\definecolor{accent-800}{HTML}{162A44}
\definecolor{accent-900}{HTML}{0F1E31}
\definecolor{accent-950}{HTML}{0A1521}

\definecolor{amber-50}{HTML}{FFFBEB}
\definecolor{amber-600}{HTML}{E67C00}   %
\definecolor{amber-800}{HTML}{92400E}

\ifdefined\GrayscaleMode
  \colorlet{accent-50}{gray-100}\colorlet{accent-100}{gray-200}
  \colorlet{accent-200}{gray-300}\colorlet{accent-300}{gray-400}
  \colorlet{accent-400}{gray-500}\colorlet{accent-500}{gray-600}
  \colorlet{accent-600}{gray-700}\colorlet{accent-700}{gray-700}
  \colorlet{accent-800}{gray-800}\colorlet{accent-900}{gray-900}
  \colorlet{accent-950}{gray-950}
  \colorlet{amber-50}{gray-100}\colorlet{amber-600}{gray-600}
  \colorlet{amber-800}{gray-800}
\fi

\colorlet{text-primary}{gray-950}
\colorlet{text-muted}{gray-600}     %
\colorlet{primary}{accent-700}      %
\colorlet{link}{accent-600}         %
\colorlet{url}{accent-600}          %
\colorlet{cite}{accent-600}         %
\colorlet{rule}{gray-300}
\colorlet{rule-strong}{accent-700}
\colorlet{highlight}{amber-600}     %
\colorlet{draft-mark}{amber-800}

\colorlet{callout-keyresult-frame}{accent-600}
\colorlet{callout-keyresult-bg}{accent-50}
\colorlet{callout-caution-frame}{amber-800}
\colorlet{callout-caution-bg}{amber-50}
\colorlet{callout-note-frame}{gray-400}
\colorlet{callout-note-bg}{gray-50}
\colorlet{callout-data-frame}{accent-400}
\colorlet{callout-data-bg}{gray-50}

\colorlet{code-bg}{gray-50}
\colorlet{code-frame}{gray-300}
\colorlet{code-keyword}{accent-700}
\colorlet{code-string}{accent-900}
\colorlet{code-comment}{gray-500}

\colorlet{pt-title}{text-primary}
\colorlet{pt-subtitle}{text-primary}
\colorlet{pt-section}{text-primary}
\colorlet{pt-section-rule}{rule}
\colorlet{pt-running-head}{text-muted}
\colorlet{pt-folio}{text-muted}
\colorlet{pt-caption}{text-primary}
\colorlet{pt-abstract-label}{text-primary}
\colorlet{pt-apparatus-label}{text-primary}   %
\colorlet{pt-draft}{draft-mark}

\definecolor{pt-inkblack}{HTML}{000000}
\colorlet{link}{pt-inkblack}
\colorlet{cite}{pt-inkblack}
\colorlet{url}{pt-inkblack}
\colorlet{primary}{pt-inkblack}
\colorlet{text-primary}{pt-inkblack}
\colorlet{pt-title}{pt-inkblack}
\colorlet{pt-section}{pt-inkblack}
\colorlet{pt-caption}{pt-inkblack}
\colorlet{pt-abstract-label}{pt-inkblack}
\colorlet{pt-apparatus-label}{pt-inkblack}

\makeatletter

\ifdefstring{\PaperLineSpacing}{double}{\doublespacing}{%
  \ifdefstring{\PaperLineSpacing}{onehalf}{\onehalfspacing}{\singlespacing}}

\newcommand{\pt@isodate}{\the\year-\two@digits\month-\two@digits\day}

\ifdefined\DraftMode
  \usepackage[switch]{lineno}
  \newcommand{\pt@draftbanner}{%
    {\normalfont\sffamily\bfseries\footnotesize\color{pt-draft}%
      DRAFT\,\textperiodcentered\,\pt@isodate\,\textperiodcentered\,Comments welcome}}
\else
  \newcommand{\pt@draftbanner}{}
\fi

\titleformat{\section}
  {\normalfont\Large\bfseries\color{pt-section}}
  {\thesection}{0.6em}{}
\titleformat{\subsection}
  {\normalfont\large\bfseries\color{pt-section}}
  {\thesubsection}{0.6em}{}
\titleformat{\subsubsection}
  {\normalfont\normalsize\bfseries\color{pt-section}}
  {\thesubsubsection}{0.6em}{}
\titlespacing*{\section}{0pt}{1.4\baselineskip}{0.6\baselineskip}
\titlespacing*{\subsection}{0pt}{1.1\baselineskip}{0.4\baselineskip}

\ifdefined\DraftMode
\else
\fi
\fancypagestyle{plain}{%
  \fancyhf{}%
  \fancyhead[C]{\pt@draftbanner}%
  \fancyfoot[C]{\color{pt-folio}\thepage}}

\theoremstyle{plain}

\theoremstyle{definition}

\theoremstyle{remark}

\makeatother

\usepackage{tcolorbox}
\tcbuselibrary{skins,breakable}

\tcbset{
  papercallout/.style={
    enhanced jigsaw,
    breakable,
    lines before break=3,
    boxrule=0.4pt,
    leftrule=2.5pt,
    arc=1pt,
    left=8pt, right=8pt, top=6pt, bottom=6pt,
    before skip=0.9\baselineskip,
    after skip=0.9\baselineskip,
    fonttitle=\normalfont\sffamily\bfseries\small,
    toptitle=5pt, bottomtitle=2pt,
    titlerule=0pt,
  },
}

\newtcolorbox{keyresult}[1][]{papercallout,
  title={Key result},
  colframe=callout-keyresult-frame, colback=callout-keyresult-bg,
  coltitle=callout-keyresult-frame, colbacktitle=callout-keyresult-bg, #1}

\newtcolorbox{caution}[1][]{papercallout,
  title={Caution},
  colframe=callout-caution-frame, colback=callout-caution-bg,
  coltitle=callout-caution-frame, colbacktitle=callout-caution-bg, #1}

\newtcolorbox{note}[1][]{papercallout,
  title={Note},
  colframe=callout-note-frame, colback=callout-note-bg,
  coltitle=callout-note-frame, colbacktitle=callout-note-bg, #1}

\newtcolorbox{databox}[1][]{papercallout,
  title={Data},
  colframe=callout-data-frame, colback=callout-data-bg,
  coltitle=callout-data-frame, colbacktitle=callout-data-bg, #1}

\usepackage{listings}

\makeatletter
\lst@AddToHook{Init}{\catcode`\#=12\relax}
\makeatother

\lstdefinestyle{papercode}{
  basicstyle=\ttfamily\small,
  keywordstyle=\color{code-keyword}\bfseries,
  commentstyle=\color{code-comment}\itshape,
  stringstyle=\color{code-string},
  showstringspaces=false,
  columns=fullflexible,
  keepspaces=true,
  tabsize=4,
  breaklines=true,
  breakatwhitespace=true,
  breakindent=1.5em,
  postbreak=\mbox{{\color{code-comment}\footnotesize$\hookrightarrow$}\space},
  frame=leftline,
  framerule=1.5pt,
  rulecolor=\color{code-frame},
  backgroundcolor=\color{code-bg},
  xleftmargin=12pt,
  framexleftmargin=6pt,
  aboveskip=0.9\baselineskip,
  belowskip=0.9\baselineskip,
  captionpos=b,
  upquote=true,
}

\lstnewenvironment{codelisting}[1][]
  {\lstset{style=papercode, #1}}
  {}

\usepackage{xspace}   %

  \usepackage{algorithm}
  \usepackage{algpseudocodex}   %

\AtBeginEnvironment{quotation}{\small}

\usepackage{xurl}       %
\usepackage[hyperfootnotes=false]{hyperref}
\usepackage{orcidlink}  %

\hypersetup{
  colorlinks=false,
  linkbordercolor={1 0 0},
  citebordercolor={0 1 0},
  urlbordercolor={0 1 1},
  pdfcreator={LaTeX},
  bookmarksnumbered=true,
  bookmarksopen=true,
}
\ifdefined\AnonMode
  \hypersetup{pdftitle={\PaperTitle}, pdfauthor={}, pdfsubject={}, pdfkeywords={}}
\else
  \hypersetup{
    pdftitle={\PaperTitle},
    pdfauthor={\PaperAuthorPlain},
    pdfsubject={\PaperAbstract},
    pdfkeywords={\PaperKeywords},
  }
\fi

\usepackage[capitalize,nameinlink]{cleveref}
\crefname{table}{Table}{Tables}
\crefname{figure}{Figure}{Figures}
\crefname{equation}{Eq.}{Eqs.}
\crefname{algorithm}{Algorithm}{Algorithms}

\begin{document}

\thispagestyle{plain}%

  \twocolumn[{%
  \begin{minipage}{\textwidth}

\begin{center}
  {\LARGE\bfseries\color{pt-title}
    \shortstack[c]{SHELF: A Synthetic Harness for\\
    Multi-Task Bibliographic Benchmarking}\par}

  \vspace{1.2em}

  \ifdefined\AnonMode
    {\large Anonymous Author(s)\par}
    \vspace{0.3em}{\small\color{text-muted}Submission under double-blind review\par}
  \else
    \PaperAuthorBlock
  \fi

  \vspace{0.8em}
  {\color{text-muted}\PaperDate\par}

\end{center}

\vspace{0.6em}

  \begin{center}
    \textbf{Abstract}
  \end{center}
  \begin{quote}
    \small\noindent\PaperAbstract
  \end{quote}

  \vspace{1.25em}
  \end{minipage}%
  }]

\PaperAuthorFootnote

\vspace{0.5em}

\ifdefined\DraftMode
  \linenumbers
\fi

\section{Introduction}
\label{sec:introduction}

Benchmarks should help users choose a model for their tasks, but that choice becomes
difficult when no benchmark covers their work or when a leaderboard score does not
predict performance under their constraints.

Libraries and archives face both of these issues. They need to know which
methods can classify, cluster, retrieve, and pair documents by subject, genre,
geography, audience, or register, and what each method requires to run.
And they need to do this across large collections, typically with limited staff
and computing budgets. Yet common benchmarks do not systematically test bibliographic
work, so libraries lack evidence that is both task-relevant and cost-aware.

At the same time, bibliographic tasks are useful measures of general fitness. A model that
cannot classify, cluster, retrieve, or pair documents by subject, genre,
geography, audience, or register likely has practical limitations that an aggregate
score may hide. In particular, a model that cannot distinguish what a document
is about from how it is written is likely to make serious mistakes when
drafting or summarizing. While success on these tasks does not establish general
fitness, failure can identify limits that matter well beyond bibliographic work.

The Massive
Text Embedding Benchmark (MTEB)
\citep{muennighoff2023mteb}, the best-known benchmark in this area, measures
model performance broadly, yet it does not include bibliographic tasks. Across
all 1{,}451 definitions in \texttt{mteb} 2.20.4, there is no Library of Congress,
Dewey, MARC, or cataloguing task.\footnote{Checked by searching the name,
description, type, and domains of every task definition on 29 August 2026.
Two names mislead. \texttt{LccSentimentClassification} uses the Leipzig
Corpora Collection, not the Library of Congress classification, and the one
task matching ``catalogue'' retrieves beverage marketing documents.}

Here enters SHELF, the Synthetic Harness for Evaluating LLM Fitness. SHELF
generates bibliographic benchmark data when real labelled text is scarce,
expensive, or contested. The software takes a labelled taxonomy, writing
specifications, and a generation budget, then writes labelled documents and
builds evaluation tasks.

The current release includes one dataset made with this system. It uses Library
of Congress vocabularies and varies subject, genre, audience, register,
geography, and length. We ask which sparse methods, encoders, and decoders can
support the resulting bibliographic tasks, although not every method supports
every task. Every benchmark score in this paper comes from this released dataset.
The only exceptions are the external validation experiments in
\Cref{sec:transfer}, which show why we do not apply its scores to other
datasets.

Library cataloguing makes this a harder problem, but also a real one. SHELF uses 21 broad
Library of Congress Classification classes and 133 genre and form labels.
Cataloguers do not always assign the same subject headings to the same work,
as the LCSHBench concordance study later quantifies
\citep{tang2026lcshbench}.

Two projects have begun to fill that gap with real catalogue records
\citep{tang2026lcshbench,dsouza2025llms4subjects}. Those records carry a
further problem, though: what a cataloguer assigned is bound up with who
catalogued it, when, and in what language. Genre and subject also move
together in natural collections, so a corpus of real records cannot tell you
whether a model reads the subject or reads the register.

SHELF therefore generates documents instead of collecting existing catalogue records.
Each document is written to a specified subject, genre, audience, register,
and other bibliographic facets. This control lets us vary facets separately,
even when they often occur together in natural collections. SHELF also varies
the writing model, prompt style, and sampling settings, so the source of each
document is measured rather than hidden.

The released dataset contains 62{,}899 documents from 25 models and records
how each document was generated. Its
factorial subset contains 18{,}345 documents from 15 models and four prompt
styles, covering all 60 model--prompt pairings. Temperature ranges from 0.6 to
1.2, and nucleus-sampling probability ranges from 0.85 to 1.0. Researchers can
use this subset when an analysis requires tighter control over the writing
model and bibliographic facets. \Cref{sec:corpus} describes the subset and
the split protocol.

All released SHELF documents were generated in or after November 2025, and every
embedding checkpoint evaluated here predates them, so the documents are out
of sample for the embedding panel. This claim does not cover the later decoder
models in \Cref{sec:decoders}. Future models may be trained on this public
release, but SHELF can then generate another dataset after those models' training cutoffs
while preserving the same labels and task design. The new documents will be
verifiably unseen, although generation after a cutoff does not remove source
bias.

The current evaluation turns those controlled documents into five kinds of
task: classification, clustering, retrieval, pair classification, and
instruction retrieval. Together, they cover subject, topic, genre, audience,
register, and geography. \Cref{tab:task-inventory} lists every task variant,
metric, and evaluated method family, while the 25 configurations span sparse
methods, early sentence encoders, and recent small and mid-size embedding
models.

The results separate easy labels from hard ones. Subject classification
reaches 0.8887, while classification across 133 genre forms reaches 0.2605.
Subject retrieval reaches 0.7104, but genre-form retrieval reaches 0.1173.
Register and audience pairs remain near chance. Sparse methods compete with
dense encoders in classification and trail them in retrieval. The separate
tasks show which formulations and label spaces current embeddings can use.

Machine-written data still require an external check, so we compare the subject
tasks with Project Gutenberg passages and LCSHBench catalogue records. SHELF
produces similar model rankings, but absolute scores do not transfer between
the corpora. We use this result to bound the benchmark claim, not to define
the benchmark: the released dataset supports model comparison and diagnosis,
not estimates of production accuracy.

In sum, this paper makes five contributions:

\begin{enumerate}
  \item A system that turns a labelled taxonomy, controlled
        writing specifications, and a generation budget into bibliographic
        benchmarking data and evaluation tasks. The released code can write a
        fresh dataset after a future training cutoff (\Cref{sec:corpus}).
  \item An evaluation dataset of 62{,}899 documents from 25
        models. Of these, 20{,}367 are specification-backed, including
        an 18{,}345-document generator-balanced factorial component, so the
        writing model is a reported axis rather than a nuisance
        (\Cref{sec:corpus}).
  \item An evaluation spanning five task formulations and six bibliographic
        facets, with stored results for 25 model configurations
        (\Cref{sec:baselines,sec:retrieval}).
  \item Measurements of task difficulty, label leakage, model headroom, and
        failed task designs (\Cref{sec:headroom,sec:surface,sec:negative}).
  \item A bounded external check of subject scores and model rankings against
        two natural corpora (\Cref{sec:transfer,sec:ranking}).
\end{enumerate}

\section{Related work}
\label{sec:related}

\subsection{Automated subject cataloguing}

Predicting a classification from a subject heading is not a new task; Frank and
Paynter described it in 2004 \citep{frank2004predicting}. Annif provides an
open-source system for automated subject indexing, and SemEval-2025 Task 5
evaluated subject tagging over roughly 100{,}000 English and German catalogue
records \citep{dsouza2025llms4subjects}.

The nearest neighbour to this paper is LCSHBench \citep{tang2026lcshbench}, released in June 2026:
22{,}346 books in 15 languages from the Harvard, Columbia, and Princeton
catalogues, admitted only when two agencies agreed. Its concordance study
over 465{,}187 works is the strongest published statement of how far
cataloguers disagree --- 93.3\% share a concept-level heading, but only
39.4\% assign identical sets.

That line of work assigns headings to real records at extreme label
scale. We measure how well a representation separates classes under a design
that fixes the confounds. The questions differ, so LCSHBench is useful here as
a test corpus rather than as a competitor. We use it that way in
\Cref{sec:transfer,sec:ranking}.

\subsection{Embedding benchmarks}

MTEB \citep{muennighoff2023mteb} and the Benchmarking Information Retrieval
collection (BEIR) \citep{thakur2021beir} have shaped the field. The
Massive Multilingual Text Embedding Benchmark (MMTEB)
\citep{enevoldsen2025mmteb} extended it to more than 500 tasks and more than
250 languages. HUME measured human performance on 16 MTEB datasets and found
only a small gap between the human average and the best model
\citep{elassadi2026hume}.
The Embedder's Dilemma compares 26 embedding models from 118M to 14B
parameters with ten language models and finds that the best aggregate scores
are close, while cost and task-level strengths differ sharply
\citep{elassadi2026dilemma}. Our smaller encoder set does not test that
frontier, a limit stated in \Cref{sec:limits}.

\subsection{Domain-specific benchmarks}

FinMTEB \citep{tang2025finmteb} finds that general rank does not reliably
predict finance-task rank. CaseHOLD \citep{zheng2021casehold} likewise builds
a domain task because existing legal tasks did not expose the intended gap.

Raji and colleagues \citep{raji2021everything} warn against benchmarks that
claim to measure broad constructs. Bean and colleagues
\citep{bean2025construct} review 445 benchmarks and find recurring weaknesses
in construct validity. Bowman and Dahl \citep{bowman2021what} include reliable
annotation among their criteria. SHELF does not yet meet that criterion, as we
acknowledge in \Cref{sec:limits}.

\subsection{Model-generated evaluation data}

Generating a corpus for evaluation is less common than generating training
examples. E5-Mistral \citep{wang2024e5mistral} trained on hundreds of thousands
of model-written examples and achieved strong benchmark results without
labelled data. PhantomWiki \citep{gong2025phantomwiki} generates a fresh,
factually consistent corpus on demand rather than using a fixed dataset.
YourBench \citep{shashidhar2025yourbench}
takes the opposite route and generates evaluation sets from a user's own
documents with language models, which makes the corpus cheap to refresh and
leaves its label semantics dependent on the generator.

Gill and colleagues \citep{gill2025lost} provide a direct warning. They wrote
model versions of two reading-comprehension datasets and found them valid by the
annotation guidelines yet easier for models than the human-written originals.
They call for direct tests of the practice. \Cref{sec:transfer} provides one,
and its transfer results support part of their criticism.

A second objection is source bias: neural retrievers rank model-written text
above equivalent human text, by a wide margin \citep{dai2024biased}. Our corpus
is written by machine throughout, so source bias may affect the reported model
rankings by an unknown amount. We return to this in \Cref{sec:limits}.

Majurski and Matuszek \citep{majurski2025grounding} show that written
benchmarks grounded in real documents can reproduce rankings from
human-curated benchmarks. \Cref{sec:ranking} tests that claim here.

\subsection{Contamination}

Synthetic generation is often defended as a guard against contamination, and
that defence is weaker than it looks. GSM1k \citep{zhang2024gsm1k} showed accuracy drops of up to
eight points on a fresh test set built to match an old one. Work on rewritten
samples \citep{yang2023rephrased} showed that a model trained on such a test set stays undetectable
by n-gram overlap, which defeats the usual decontamination check.

We claim only what timing buys. The released SHELF documents postdate the
evaluated embedding checkpoints, and the system can generate new records after
a future model's training cutoff. These properties protect against direct
overlap, but neither removes source bias. The unlikely combinations in the
factorial design do something different: they test whether a method can
separate facets that usually move together.

\subsection{Instruction-conditioned retrieval}

Instruction-following retrieval is covered by FollowIR
\citep{weller2025followir} and NevIR \citep{weller2024nevir}. FollowIR finds
that existing retrieval models often fail to use detailed instructions; many
instruction-aware retrieval models score below zero on its paired measure.

NevIR finds bi-encoder and sparse systems at or below random performance. CoDeR
\citep{yin2026coder} measures the fraction of queries whose top-$k$ results
contain evidence that violates a stated constraint.

Together, this work leaves a clear gap. Cataloguing benchmarks use real records
at large label scale, while general embedding benchmarks cover many tasks but
not bibliographic work. Model-generated benchmarks show what controlled data
can provide, but also why source bias and contamination claims require care.
SHELF joins these lines without replacing them. It uses controlled generation
to separate bibliographic facets, compares sparse methods and encoders across
compatible tasks, and adds a subject-only decoder comparison. It then checks
subject rankings against natural corpora, so those comparisons bound what the
synthetic scores can mean. That combination---control inside SHELF and
calibration outside it---is the basis of the evaluation that follows.

\section{Corpus and evaluation protocol}
\label{sec:corpus}

\subsection{Corpus composition and splits}

The \texttt{all} subset released on Hugging Face contains 62{,}899 documents and is
the corpus used for the main classification, retrieval, clustering, and
headroom results. It combines 42{,}532 documents generated in late 2025 with
20{,}367 documents generated in August 2026. The two components share the
same principal label spaces but differ in provenance. The original records predate stored
specifications and retain their stratified document-level 60/20/20 splits.
The later records carry specifications and use stratified 60/20/20 splits
grouped by \texttt{spec\_id}. The pooled configuration preserves each
document's source split. Consequently, the specification-level leakage
guarantee applies to the later component, not to the pooled corpus as a
whole. All reported runs use seed 42 and body text only.

The generator-balanced \texttt{v0\_4\_core} component contains 18{,}345 of
the later documents. Claims that depend on factorial control or generator
balance name that component explicitly; claims reported on \texttt{all} use
the full 62{,}899-document corpus.

Each such document begins as a \emph{specification}: a subject class, a genre and
form, a set of topics, an audience, a register, a geographic region, and a
target length. The specification is hashed, and that hash is the document's
\texttt{spec\_id}.

Two properties follow from this. Every writing model receives the same specification,
so a difference between models is a difference in writing rather than in
assignment. And every document that shares a \texttt{spec\_id} is a separate
version of one brief, so those versions can be kept together when the corpus
is split.

Splitting at the document level puts near-duplicate versions of the same brief
on both sides of the split. In a pilot corpus of 600 specifications, each
written by eight generators, 598 specifications straddled a split that way. Grouping on
\texttt{spec\_id} leaves none straddling.

\subsection{Independence by construction}

The facets are drawn independently. A natural corpus cannot guarantee this,
because subject, genre, and register are typically correlated. SHELF instead
contains combinations unlikely to occur in most collections: a technical
puzzle set about agricultural conservation and quantum mechanics, professional
sermons about networks and biotechnology, and casual maps about ecology and
geology. These are real
examples from the generator-balanced slice \texttt{v0\_4\_core}, not invented
illustrations.

That slice is also balanced across writing models. The measured association
between writing model and label is negligible: Cram\'er's $V = 0.016$
for subject class and $0.027$ for genre category. Across the wider v0.4 generation,
which adds single-generator supplementary
documents, subject-class association rises to $0.037$, and any claim that requires
independent generation should use the \texttt{v0\_4\_core} subset. The implication
is that if genre and subject usually move together, a model that reads only the
register will score well on a subject task, and the benchmark cannot tell which it read.

\subsection{Scale and generation models}

The balanced component holds documents from 15 current-generation
models across 11 laboratories --- Anthropic, OpenAI, Google, Alibaba,
DeepSeek, Zhipu, Moonshot, MiniMax, Meta, Mistral, and xAI --- with the
largest single model at 9.24\%. For comparison, E5-Mistral used one model and
Cosmopedia \citep{benallal2024cosmopedia} used one model.

The aggregate corpus is not generator-balanced: its largest writing model
supplies 47.7\% of the documents.

Thus scale describes the aggregate corpus, while generator balance describes
the factorial component. We name the component when a result depends on that
balance.

A third slice, \texttt{transfer\_gutenberg}, holds 3{,}016 Project Gutenberg
passages. These passages are human-written and human-catalogued. We use them
only to test whether model rankings from SHELF carry over to natural text. We
do not include them in the main corpus, because doing so would erase the
distinction that the transfer test is meant to measure.

\begin{table}[t]
  \centering
  \scriptsize
  \setlength{\tabcolsep}{2pt}
  \begin{tabular}{@{}lrrrrrrrr@{}}
    \toprule
    Slice & Docs & Mean & Median & SD & Min & 5th & 95th & Max \\
    \midrule
    SHELF, \texttt{all}         & 62{,}899 & 612 & 326 & 782 & 1   & 34  & 2{,}369 & 6{,}203 \\
    SHELF, \texttt{v0\_4\_core} & 18{,}345 & 548 & 341 & 685 & 5   & 32  & 1{,}916 & 4{,}451 \\
    Gutenberg transfer          &  3{,}016 & 480 & 449 &  92 & 184 & 401 &     671 &     896 \\
    LCSHBench transfer          &  4{,}924 & 140 &  95 & 146 & 2   & 19  &     392 & 3{,}388 \\
    \bottomrule
  \end{tabular}
  \caption{Words per document, counted on whitespace over the body text of
  every document in each slice. 5th and 95th are percentiles.}
  \label{tab:lengths}
\end{table}

\Cref{tab:lengths} summarises document length. SHELF is right-skewed by
design: the eight target-length categories run from a median of 18 words for
micro to 2{,}985 for extended, and the short and medium categories together
hold nearly half the corpus. The shortest documents are one-line micro-length
forms such as prayers and greeting cards; 28 documents have ten words or
fewer. The three corpora also differ in shape. Gutenberg passages are
fixed-size extracts that cluster between 400 and 700 words, while LCSHBench
records are catalogue metadata with a median of 95 words. The 200-word
truncation in \Cref{sec:transfer} controls for that difference.

\subsection{Quality control}

Every document was checked before it entered the released corpus. The checks
required a parseable response in the requested language and length range. They
also rejected refusals, near-duplicates, documents that missed their assigned
topics, and documents that stated their own catalogue labels.

The first version of the label check flagged 15 to 20\% of the documents, but
most of those flags were false. It treated any use of a label as self-labelling,
so ordinary phrases in the text could trigger it. We narrowed the check to
places where a label was presented as a label, such as a heading, a term
followed by a colon, or a taxonomy code in parentheses. The revised check
flagged 2.48\% of the earlier corpus and less than 0.5\% of the newer corpus.

One generation failure was more serious. For one model, 14.6\% of calls
returned reasoning tokens but no document, mostly on requests for short text.
We increased the output-token allowance and removed every affected record.
Both corrections changed which documents entered the release. Researchers
who reproduce this work should check how each provider handles reasoning
tokens, thinking budgets, and output-token limits, especially when requesting
short documents.

\subsection{Evaluation protocol}

\Cref{tab:task-inventory} is the complete task inventory for this paper. The
first five rows are released SHELF benchmark formulations. The final row is a
separate decoder comparison on one classification task.

\begin{table*}[t]
  \centering
  \scriptsize
  \setlength{\tabcolsep}{3pt}
  \begin{tabularx}{\textwidth}{@{}p{0.13\textwidth}p{0.20\textwidth}Xp{0.09\textwidth}p{0.17\textwidth}@{}}
    \toprule
    Formulation & Short description & Task variants & Metric & Evaluated methods \\
    \midrule
    Classification
      & predict one label
      & subject; genre category; genre form; register
      & macro-$F_1$
      & TF; TF-IDF; encoders \\
    Clustering
      & recover label groups
      & subject; genre form; register; geography
      & ARI
      & TF; TF-IDF; encoders \\
    Retrieval
      & rank documents for a label
      & subject; genre category; genre form
      & nDCG@10
      & TF-IDF; BM25; encoders \\
    Pair classification
      & score a document pair
      & same subject; same topic; topic overlap; same genre form; same register; same audience
      & AUC
      & TF-IDF; BM25; encoders \\
    \makecell[l]{Instruction\\retrieval}
      & match one facet and differ on another
      & same subject/different form; same topic/different subject; same form/different subject; same audience/different register
      & nDCG@10
      & TF-IDF; BM25; encoders \\
    Decoder reference
      & predict a label without SHELF examples
      & subject only
      & macro-$F_1$
      & four decoder models \\
    \bottomrule
  \end{tabularx}
  \caption{Tasks evaluated in this paper. Not every method supports every
  formulation. ARI is the adjusted Rand index.}
  \label{tab:task-inventory}
\end{table*}

Panel size changes with the formulation. Classification and clustering use
25 configurations. Retrieval also uses 25, replacing term frequency with
BM25. Instruction retrieval uses the same 25 retrieval-compatible methods.
The released pair configurations come from an earlier 21-model sweep. The
cross-corpus analysis counts 24 distinct embedding models, or 25 for pairs,
after accounting for task-specific methods. The masking
analysis uses the 21 models for which every required arm was run.

The model comparison uses frozen encoders. For classification, we fit
scikit-learn 1.8.0 logistic regression on the training embeddings, with
class-balanced weights, a maximum of 1{,}000 iterations, seed 42, and otherwise
default settings. We report macro-$F_1$ on the test split. For retrieval, test documents are queries and
the training and validation documents form the search corpus; rankings use
cosine similarity and are scored with nDCG@10. For clustering, we apply
$k$-means to normalised test embeddings with $k$ set to the number of labels
and report adjusted Rand index. Pair tasks use cosine similarity and report
AUC, which does not require choosing a decision threshold. Exact checkpoint
names, package versions, corpus checksums, and per-task result files are in
the evaluation repository at \url{https://github.com/mjbommar/shelf-benchmark}.

\section{Benchmark results}
\label{sec:baselines}

We evaluate 25 configurations on SHELF classification and clustering tasks.
The panel includes lexical methods and encoders from 22M to 596M parameters.
Four encoders were released in 2025; one appears at two context lengths, so
they account for five configurations. We report macro-$F_1$ for classification
and adjusted Rand index for clustering. An adjusted Rand index near zero means
that the clusters are no better than a random partition. We keep the two
metrics separate because they measure different tasks.

\subsection{Evaluated methods}

The lexical baselines are term frequency and TF-IDF, both reduced with
truncated SVD. BM25 \citep{robertson2009bm25} is included only on tasks that
support its scoring interface.

The encoder panel begins with BERT \citep{devlin2019bert}, RoBERTa
\citep{liu2019roberta}, and DistilBERT \citep{sanh2019distilbert}, used as
frozen backbones with mean pooling. It also
includes MiniLM \citep{wang2020minilm}, MPNet \citep{song2020mpnet}, the BGE
\citep{xiao2024bge}, E5 \citep{wang2022e5}, and GTE \citep{li2023gte}
families, GTR-T5 \citep{ni2022gtr}, and Instructor
\citep{su2023instructor}. This range lets us compare early sentence encoders
with models trained more directly for embedding and retrieval.

The panel also includes two masked-language models from the OGBert family,
trained on the OpenGloss synthetic dictionary corpus
\citep{bommarito2025opengloss}. We label the published
110M model and the smaller 36M model as OGBert 110M and OGBert 36M. Their
model cards document the training data and checkpoints.\footnote{\url{https://huggingface.co/mjbommar/ogbert-110m-base} and
\url{https://huggingface.co/mjbommar/ogbert-v1-mlm}.}

Four encoders in the panel were released in 2025: Granite-small-r2 at 48M
parameters, GTE-ModernBERT at 149M \citep{warner2024modernbert},
EmbeddingGemma at 308M \citep{vera2025embgemma}, and Qwen3-Embedding at 596M
\citep{zhang2025qwen3emb}. We run GTE-ModernBERT with context limits of
2{,}048 and 8{,}192 tokens to measure the effect of that limit
(\Cref{sec:limits}). The largest of these newer encoders has 596M parameters;
the largest older encoder in the panel has 335M. Thus, the panel remains a
comparison among small and mid-size encoders; it does not compare them with
frontier-scale models.

We do not add the query or document prefixes recommended by individual model
cards. Every encoder receives the same unprompted input. This keeps the
protocol consistent, but it may understate models that depend on those
prefixes and may affect their order in the tables.

\subsection{Classification}

\begin{table*}[!t]
  \centering
  \small
  \begin{tabular}{lcccc}
    \toprule
    Model & Subject (21) & Category (14) & Genre form (133) & Register (8) \\
    \midrule
    EmbeddingGemma-300M$^\dagger$ & \textbf{0.8887} & \textbf{0.7988} & \textbf{0.2605} & 0.6013 \\
    Qwen3-Embedding-0.6B$^\dagger$ & 0.8823 & 0.7156 & 0.1933 & 0.5347 \\
    TF-IDF+SVD        & 0.8686 & 0.7342 & 0.1735 & \textbf{0.6358} \\
    BGE-large         & 0.8663 & 0.7395 & 0.2069 & 0.5525 \\
    GTE-ModernBERT-8k$^\dagger$ & 0.8655 & 0.7589 & 0.2269 & 0.5291 \\
    GTE-ModernBERT-2k$^\dagger$ & 0.8646 & 0.7575 & 0.2281 & 0.5301 \\
    E5-large          & 0.8512 & 0.7521 & 0.2000 & 0.5882 \\
    Granite-small-r2$^\dagger$ & 0.8510 & 0.6713 & 0.1710 & 0.4505 \\
    E5-base           & 0.8492 & 0.7461 & 0.2106 & 0.5658 \\
    GTE-base          & 0.8469 & 0.7129 & 0.1837 & 0.4964 \\
    BGE-base          & 0.8445 & 0.7179 & 0.2022 & 0.5329 \\
    MPNet             & 0.8428 & 0.6506 & 0.1418 & 0.4629 \\
    GTE-small         & 0.8386 & 0.7072 & 0.1762 & 0.4608 \\
    BGE-small         & 0.8342 & 0.7073 & 0.1960 & 0.4735 \\
    GTR-T5-large      & 0.8300 & 0.7056 & 0.1634 & 0.5158 \\
    E5-small          & 0.8249 & 0.7176 & 0.1793 & 0.5243 \\
    Instructor-base   & 0.8204 & 0.7194 & 0.1786 & 0.5173 \\
    GTR-T5-base       & 0.8079 & 0.6582 & 0.1467 & 0.4785 \\
    TF+SVD            & 0.8048 & 0.6450 & 0.1288 & 0.5103 \\
    MiniLM-L6         & 0.8015 & 0.6228 & 0.1476 & 0.4374 \\
    BERT              & 0.7928 & 0.6913 & 0.1481 & 0.5983 \\
    DistilBERT        & 0.7811 & 0.6782 & 0.1302 & 0.5784 \\
    OGBert 110M       & 0.7449 & 0.6178 & 0.0951 & 0.5281 \\
    OGBert 36M        & 0.6913 & 0.5605 & 0.0820 & 0.4846 \\
    RoBERTa           & 0.6715 & 0.6142 & 0.0734 & 0.5200 \\
    \bottomrule
  \end{tabular}
  \caption{Classification, macro-$F_1$, on the pooled corpus. Ordered by the
  21-class subject task. Label counts are in the column heads. Rows marked
  $\dagger$ are 2025 encoders; \texttt{GTE-ModernBERT} appears twice, at
  2{,}048 and 8{,}192 tokens on identical weights.}
  \label{tab:baselines-cls}
\end{table*}

The word-frequency baseline is competitive. TF-IDF leads the register task,
comes third on the 21-class subject task, and
sits mid-table on genre form. BEIR established that BM25 is hard to displace
in zero-shot retrieval \citep{thakur2021beir}, and FinMTEB reported
bag-of-words beating every dense model on financial semantic similarity
\citep{tang2025finmteb}. Our results reproduce that pattern.

Model size buys less than expected. Models at 335M parameters and
above occupy ranks two, four, seven, and fifteen on the subject task. Three sit behind
a sparse baseline with no parameters to speak of.

\subsection{Clustering}

\begin{table}[t]
  \centering
  \scriptsize
  \setlength{\tabcolsep}{3pt}
  \begin{tabular}{@{}llrr@{}}
    \toprule
    Task & Best model & Best & Median \\
    \midrule
    Subject   & Qwen3-Embedding-0.6B & \textbf{0.5152} & 0.3439 \\
    Genre form & EmbeddingGemma-300M & 0.0921 & 0.0484 \\
    Register  & RoBERTa              & 0.0678 & 0.0110 \\
    Geography & GTR-T5-base          & 0.0068 & 0.0025 \\
    \bottomrule
  \end{tabular}
  \caption{Best model, best score, and median adjusted Rand index for each
  clustering task over 25 configurations on the pooled corpus. Zero is chance. These are
  single-run values at seed 42. The rank-agreement analysis in
  \Cref{sec:ranking} uses each model's median over five seeds instead,
  because clustering is seed-sensitive: the subject task's best and median
  become 0.5247 and 0.3456 under that measure rather than 0.5152 and 0.3439.}
  \label{tab:baselines-clu}
\end{table}

Geography and register do not separate the current models: their best scores
are 0.0068 and 0.0678. Genre form separates them only weakly, with a best
score of 0.0921. Subject clustering is the only useful leaderboard among the
four, reaching 0.5152 and spreading the field.

\subsection{Retrieval, pair classification, and instructions}
\label{sec:retrieval}

The remaining released formulations rank or score documents rather than
assign a label. \Cref{tab:retrieval,tab:pairs,tab:instruction} report them
in turn.

\subsubsection{Retrieval}

\begin{table}[t]
  \centering
  \scriptsize
  \setlength{\tabcolsep}{3pt}
  \begin{tabular}{@{}llrrr@{}}
    \toprule
    Task & Best model & Best & Median & BM25 \\
    \midrule
    Subject    & Qwen3-Embed-0.6B & \textbf{0.7104} & 0.5988 & 0.4533 \\
    Genre form & E5-base          & 0.1173 & 0.0947 & 0.0585 \\
    Category   & RoBERTa          & 0.4968 & 0.4247 & 0.2704 \\
    \bottomrule
  \end{tabular}
  \caption{Best model, best score, panel median, and BM25 nDCG@10 for each
  retrieval task over 25 configurations on the pooled corpus.
  ``Qwen3-Embed-0.6B'' abbreviates Qwen3-Embedding-0.6B.}
  \label{tab:retrieval}
\end{table}

Genre form is as hard to retrieve as it is to classify: the best model
reaches 0.1173, against 0.7104 for subject. Classification and retrieval both
identify genre form as the harder label space.

RoBERTa leads category retrieval despite ranking last on the subject and
genre-form classification tasks. Because raw RoBERTa lacks sentence-level similarity
training, we treat this as a task-specific result rather than evidence of
generally stronger representations \citep{reimers2019sbert}.

BM25 ranks 23rd, 24th and 25th of 25 on the three retrieval tasks. TF-IDF
leads the register classification task and ranks third on subject
(\Cref{tab:baselines-cls}), so sparse strength is task-specific.

\subsubsection{Pair classification}

Each pair task is a balanced binary decision: 2{,}000 positive and 2{,}000
negative pairs. We report AUC because it measures how well cosine similarity
orders positive above negative pairs without selecting a threshold on the
test set. BM25 has no embedding, so its pair score is the mean of the BM25
score of each document used as a query against the other.
\Cref{tab:pairs} summarises the six tasks.

\begin{table}[t]
  \centering
  \begin{tabular}{lcc}
    \toprule
    Task & Best & Median \\
    \midrule
    Same subject    & \textbf{0.8474} & 0.6925 \\
    Same topic      & 0.8163 & 0.7054 \\
    Topic overlap   & 0.8282 & \textbf{0.7133} \\
    Same genre form & 0.6579 & 0.6174 \\
    Same register   & 0.5718 & 0.5329 \\
    Same audience   & 0.5519 & 0.5331 \\
    \bottomrule
  \end{tabular}
  \caption{Pair classification AUC over 21 models, including BM25, on the
  released pair configurations, which use the original corpus component
  described in \Cref{sec:corpus} rather than the aggregate corpus used
  elsewhere in this section. Chance is 0.5.}
  \label{tab:pairs}
\end{table}

Models distinguish pairs by subject, topic, and topic overlap, with median AUC
from 0.69 to 0.71. They distinguish genre form less reliably, at 0.62, while
register and audience remain close to chance.

Four thousand pairs are not four thousand independent observations, because they are built
from far fewer documents and each document can appear in several pairs. We
therefore bootstrap documents rather than individual pairs. When a document
is resampled, we include every pair that contains it. Rebuilding the same-subject task on the pooled
corpus and bootstrapping that way gives 0.8309 with a 95\% document-clustered
interval of $[0.8093, 0.8511]$ for the best model.

The 0.8474 in \Cref{tab:pairs} comes from the released pair configurations, which are
mined from the original corpus component; the 0.8309 is the same task mined
from the aggregate corpus.
The two corpora produce different pair sets and different scores. The table
uses the released configurations because readers can download them.

\subsubsection{Instruction retrieval}

\begin{table}[t]
  \centering
  \small
  \setlength{\tabcolsep}{4pt}
  \begin{tabular}{lcc}
    \toprule
    Instruction & Best & Median \\
    \midrule
    Same subject, different form     & \textbf{0.5496} & 0.4724 \\
    Same topic, different subject    & 0.2030 & 0.1771 \\
    Same form, different subject     & 0.0764 & 0.0439 \\
    Same audience, different register& 0.0680 & 0.0507 \\
    \bottomrule
  \end{tabular}
  \caption{Instruction-conditioned retrieval, nDCG@10. Each task asks for
  documents matching one facet while differing on another.}
  \label{tab:instruction}
\end{table}

Instructions anchored on subject are followed to a useful degree; instructions
anchored on form or audience are not. This resembles FollowIR, where many
instruction-aware retrieval models scored below zero on the paired measure.
The present task does not isolate instruction following from the strength of
the underlying label signal.

\subsection{Task difficulty and headroom}
\label{sec:headroom}

The subject task has 21 classes, while the genre-form task has 133. We score
all 25 configurations on both tasks, using the same corpus and documents. This
lets us ask two separate questions: how much harder is genre form, and does it
change the order of the models? The two relevant columns appear in
\Cref{tab:baselines-cls}.

\subsubsection{Differences between tasks}

Genre form is much harder. The best score falls from 0.8887 on subject to
0.2605 on genre form. Every model loses between 0.60 and 0.70 points, and the
mean falls from 0.8226 to 0.1698.

This is not an artifact of unseen classes. All 133 forms appear in every
split, the rarest form in the test split has 25 documents, and the same
documents carry both labels.

\subsubsection{Agreement in model rankings}

The difficulty changes, but the model order mostly does not. Across the 24
distinct models, the two tasks have a rank correlation of 0.815, with a
bootstrap interval of $[0.58, 0.94]$.

A practitioner choosing among these models gets roughly the same order from
either task.

\subsubsection{Interpreting the genre-form task}

Genre form is useful because it remains difficult, not because it produces a
new model order. The top twelve subject scores span only 0.046, while every
genre-form score remains low. A nearly saturated subject task can hide how
much work remains; the genre-form task cannot. Its best score is only 0.2605.

An early partial run gave a different impression. The first nine completed
models had a rank correlation of 0.450, but that group was small and skewed
toward cheap models; none had more than 109M parameters. After all models
finished, the correlation rose to 0.815. The partial estimate had an interval
that included zero and did not support a claim. We did not archive that run,
so we do not report its interval.

\subsection{Zero-shot decoder comparison}
\label{sec:decoders}

Cataloguers have not annotated a SHELF sample, so the benchmark has no human
agreement rate or expert ceiling. We add a separate comparison with four
decoder language models, which generate text one token at a time. We ask each
model to assign a subject class to each of the 12,504 test documents.

These results are a separate zero-shot reference, not part of the
25-configuration main panel. They do not enter the rank analysis in \Cref{sec:ranking}.
The decoder models postdate the original public SHELF release and may have
encountered it during training. Their results do not carry the temporal
out-of-sample guarantee that applies to the embedding panel.

\subsubsection{Classification results}

\begin{table*}[!t]
  \centering
  \begin{tabular}{llcc}
    \toprule
    Model & Parameters & Macro-$F_1$ & Accuracy \\
    \midrule
    Qwen3.5-0.8B          & 0.87B        & 0.3822 & 0.3912 \\
    Gemma-4-E2B           & 5.12B        & 0.4557 & 0.4959 \\
    Qwen3.5-2B            & 2.27B        & 0.5075 & 0.5263 \\
    GPT-5.6-luna          & undisclosed  & \textbf{0.5860} & 0.6068 \\
    \midrule
    Best encoder, supervised & 0.30B     & \textbf{0.8887} & \\
    \bottomrule
  \end{tabular}
  \caption{Decoders on 21-way subject classification, zero-shot, over the same
  12,504 test documents as \Cref{tab:baselines-cls}. The encoder row is a
  \emph{supervised} linear probe fitted on 37,795 labelled documents; the
  decoders see no labelled SHELF examples. The rows therefore do not compare
  representation quality: one method has a fitted classifier and the other
  does not.}
  \label{tab:decoders}
\end{table*}

The class names contain useful information. A uniform guess over
21 classes has an expected macro-$F_1$ of 0.048. Every decoder scores well
above that value and predicts all 21 classes. Without labelled SHELF examples,
the models can connect document text with the meanings of the class names.
This result does not show whether the models encountered Library of Congress
labels during pretraining.

The supervised probe scores higher, but the comparison is not like for like.
The best decoder
reaches 0.5860 against 0.8887 for the best encoder, and that encoder has
0.30B parameters. The encoder arm fits a
classifier on 37,795 labelled documents, while the decoder arm receives no
labelled SHELF examples. In this experiment, the small supervised encoder
scores higher than all four zero-shot decoders. The decoder results give a
reference for the tested models and prompts, not a general zero-shot floor.

\subsubsection{Prompt and generator effects}

Prompt selection affects the scores. We selected the prompt and
context limit for each model on the validation split. Across the tested
settings, macro-$F_1$ changed by more than 0.4 for one model, and different
prompts worked best for different models. Using the chat template improved
mean macro-$F_1$ by a factor of two to three over raw completion. The table
reports each model's best validation setting. The repository contains the
full validation grid.

We find no consistent own-family advantage.
Models in the GPT and OpenAI generator groups wrote 68.4\% of the test split,
so a GPT judge is largely reading text from its own family. Own-family accuracy differs from
other-family accuracy by $+0.0006$, $-0.0533$, $+0.0065$, and $+0.0271$ for
the four judges. The direction is not consistent. The pooled test split has
twelve stored generator groups. Legacy metadata separates \texttt{gpt} and
\texttt{openai}, so these groups are not the eleven laboratories counted for
the balanced component in \Cref{sec:corpus}. Across the twelve groups, the six pairwise rank correlations between judges range from 0.622
to 0.902. The judges broadly agree about which groups of documents are harder,
but these descriptive results do not identify the cause.

\subsubsection{Measured inference rates}
\label{sec:timing}

Accuracy alone does not determine which method to deploy. We measured local
inference on the first 1{,}000 test documents. The neural inputs were capped at
no more than 512 tokens from each model's tokenizer. The lexical model read
the whole document. We discarded a warm-up pass and excluded model loading and
classifier fitting. The rates come from one run, so they describe this harness
and machine rather than each model's best possible throughput.

\begin{table*}[!t]
  \centering
  \begin{tabular}{llccc}
    \toprule
    Arm & Model & Macro-$F_1$ & Docs/s & Amortized ms/doc \\
    \midrule
    Lexical, CPU     & TF-IDF + logistic       & 0.8686 & \textbf{2270.9} & 0.44 \\
    \midrule
    Encoder + probe  & MiniLM-L6               & 0.8015 & 715.9 & 1.40 \\
    Encoder + probe  & BGE-small               & 0.8342 & 379.3 & 2.64 \\
    Encoder + probe  & EmbeddingGemma-300M     & \textbf{0.8887} & 119.2 & 8.39 \\
    \midrule
    Decoder, 0-shot  & Qwen3.5-0.8B            & 0.3822 & 11.5 & 86.92 \\
    Decoder, 0-shot  & Gemma-4-E2B             & 0.4557 & 20.3 & 49.24 \\
    Decoder, 0-shot  & Qwen3.5-2B              & 0.5075 & 11.2 & 89.72 \\
    Decoder, 0-shot  & GPT-5.6-luna (API)      & 0.5860 & 9.4 & 106 \\
    \bottomrule
  \end{tabular}
  \caption{Reported test macro-$F_1$ and observed processing rates. Accuracy
  uses the full test evaluation. Local rates use 1{,}000 documents on one host:
  TF-IDF on CPU, encoders on an RTX 4070 Ti SUPER in batches of 64, and local
  decoders on that GPU one document at a time. Rate measurements cap neural
  inputs at 512 model-specific tokens; TF-IDF read the whole document. The
  macro-$F_1$ values use each method's frozen test configuration, which uses
  2{,}048 tokens for Qwen3.5-0.8B and whole documents for Qwen3.5-2B. The API
  row uses concurrent requests over the full test run. The final column
  is elapsed time divided by document count, not single-request latency.}
  \label{tab:timing}
\end{table*}

The lexical baseline is strong on this task. TF-IDF with a
logistic head processes 2{,}271 documents per second on CPU and scores 0.8686.
It trails the best encoder by 0.0201 macro-$F_1$ and has nineteen times its
measured processing rate. For supervised subject classification, TF-IDF is a
practical baseline when a system does not need reusable embeddings for other
tasks.

The encoders show an accuracy--rate tradeoff.
EmbeddingGemma-300M improves macro-$F_1$ by 0.0872 over MiniLM-L6. Its measured
processing rate is about one sixth as high. The table reports the tradeoff but
does not assign a dollar or energy cost to it.

Parameter count alone does not determine processing rate.
Gemma-4-E2B has 5.12B weights but processes 20.3 documents per second in this
harness. Qwen3.5-0.8B has 0.87B weights and processes 11.5. Architecture,
implementation, and batching also affect processing rate. This comparison
does not isolate their effects or predict rates on another system.

Input length changes the rate. Qwen3.5-0.8B processes 9.6
documents per second at its frozen 2{,}048-token limit, compared with 11.5 at
the 512-token cap. Qwen3.5-2B processes 7.2 whole documents per second,
compared with 11.2 at the cap. The macro-$F_1$ values in the table come from
the frozen configurations. The 512-token rates for these two models therefore
do not describe the exact configurations that produced those scores.

\section{Benchmark diagnostics}
\label{sec:surface}

The aggregate scores do not show what the models read or where a task fails.
We test the most important sources of misleading performance before turning
to external validation.

\subsection{Surface label signal}

SHELF documents often repeat words from their own labels. This surface signal
can make a task easier without making the underlying representation better.

\subsubsection{Measuring label terms}

Count how often a document contains the exact name of its own label. Reported
alone, the number has no clear meaning. A document about birds will contain the
word ``birds'', and it should; a corpus scoring zero would be a strange corpus.

The number becomes readable when a natural corpus is measured the same way.

\begin{table}[t]
  \centering
  \begin{tabular}{lcc}
    \toprule
    Corpus & Full text & First 200 words \\
    \midrule
    Gutenberg (natural) & 9.0\% & 6.1\% \\
    SHELF               & 23.4\% & 19.1\% \\
    \bottomrule
  \end{tabular}
  \caption{How often the subject-class name appears in its own document.
  Both corpora are running prose, and the right column cuts every
  document to its first 200 words so length cannot drive the comparison.
  Counted exhaustively over all 65{,}915 documents, so the figures are exact
  rather than sampled.}
  \label{tab:surface}
\end{table}

Real documents do name their own subjects, about 6\% of the time at matched
length. SHELF does it about three times as often. That excess is consistent
with the higher within-corpus score, but this count does not estimate how much
of the score difference it causes.

\subsubsection{Removing the label terms}

The count alone cannot tell us how much these terms affect a score, so we run
two deletion tests. The \emph{masked} arm removes every occurrence of a
document's own label terms. Those terms are broader than the subject-class
name counted in \Cref{tab:surface}: they cover the subject-class name, the
document's topic terms, and its genre form and category names, each with a
frozen variant list of the plural and every token of five or more characters. The \emph{sham} arm removes the same number of
random tokens from the same document. Comparing the two separates the effect
of removing label words from the effect of shortening the text.

Every document loses the same token count in the masked and sham arms. We
check the count per document, not only in the mean. Masking can affect only
documents that contain their own label terms. That is 86.3\% of SHELF, 52.3\%
of Gutenberg, and 24.1\% of LCSHBench, so the test has less power on LCSHBench
than on SHELF.

The number of removed tokens also differs because the corpora carry different metadata.
Masking removes an average of 9.13 tokens from SHELF, 1.89 from Gutenberg,
and 0.63 from LCSHBench. We therefore compare each masked score with its own
sham rather than comparing masked scores across corpora.

\begin{table*}[t]
  \centering
  \begin{tabular}{lccc}
    \toprule
    Corpus & Unmasked & Masked & Sham \\
    \midrule
    SHELF     & 0.7978 & 0.7641 ($-0.0337$) & 0.7974 ($-0.0004$) \\
    Gutenberg & 0.4318 & 0.4203 ($-0.0115$) & 0.4325 ($+0.0007$) \\
    LCSHBench & 0.5248 & 0.5012 ($-0.0235$) & 0.5264 ($+0.0016$) \\
    \bottomrule
  \end{tabular}
  \caption{Subject classification, macro-$F_1$, averaged over the 21 distinct
  models measured in the masking arms. Masking deletes the label terms; the sham deletes as many random
  tokens from the same documents. The sham column is the control: what a
  document loses by being shortened at all.}
  \label{tab:masking}
\end{table*}

Removing label terms costs about 3.4 macro-$F_1$ points on SHELF and 1.2 on
Gutenberg, while the sham moves neither score materially. Retrieval behaves
the same way, at $-0.0396$ masked against $+0.0008$ sham on SHELF. Label-term
availability therefore contributes measurably to SHELF's absolute scores,
more than it does in Gutenberg.

Masking changes the rankings only slightly. Correlation between masked and
unmasked model order is 0.987 on SHELF, 0.988 on Gutenberg, and 0.983 on
LCSHBench. Removing random tokens changes none of the orders: the sham
correlation is 1.000 $[1.00, 1.00]$. On SHELF, the difference between the
masked and sham correlations is $-0.013$, with an interval of
$[-0.063, +0.000]$. The cross-corpus rankings also remain similar after
masking. SHELF's agreement is 0.831 with Gutenberg and 0.748 with LCSHBench,
compared with 0.870 and 0.765 before masking.

Label-term availability raises scores but has little effect on model order.

\subsubsection{Changes between generations}

SHELF exists in two generations built with different prompts and gates. Their
difference can be measured, but it cannot be assigned to one change in the
pipeline.

\begin{table}[t]
  \centering
  \begin{tabular}{lccc}
    \toprule
    Generation & Subject class & Topics & Genre form \\
    \midrule
    First  & 21.8\% & 76.7\% & 7.3\% \\
    Second & \textbf{13.4\%} & \textbf{45.0\%} & \textbf{1.4\%} \\
    \bottomrule
  \end{tabular}
  \caption{Label names appearing in their own documents, by generation, every
  document truncated to 200 words.}
  \label{tab:qc}
\end{table}

Genre-form repetition fell by 81\% and is now rare. Topic repetition fell less
and remains common.

\subsubsection{Topic leakage and mitigation}

The cause is in the prompt. Genre and subject reach the writing model as
\emph{descriptions} --- ``spatial representations, geographic
visualizations'' rather than ``Maps''. Topics reach it as the exact strings.
This difference accounts for the pattern in \Cref{tab:qc}.

We tried describing topics too and rejected it
before spending anything, because only 382 of 1{,}983 topic descriptions come
from real scope notes. Another 1{,}546 are built from position in the
subject tree and pick the wrong sense: ``Information'' resolves to a topic
within criminal procedure, ``Cloud computing'' to distributed data
processing, ``Security'' to investments. Writing from those would cut the leak
by writing about the wrong subject.

What does work is an instruction. Telling the model to develop the subject
without using the topic words as labels cuts the leak from 82.2\% to 4.4\% on
one model and from 61.5\% to 23.1\% on another.

Removing the topic words has a cost, though. We asked a blind judge, one that
did not know which prompt had produced each document, whether the document
still covered its assigned topics: 543 judgements across two generators. With
the plain instruction, topic coverage fell by 11.7 points, with an interval
from $-19.5$ to $-3.4$. When the instruction also supplied a trustworthy gloss
of the topic, coverage fell by 6.5 points, with an interval from $-13.7$ to
$+1.2$. The plain instruction therefore trades leakage for coverage. The gloss
version may avoid that trade, but its interval includes zero, so we cannot say.
The released generation keeps the original prompt.

We did not measure coverage by cosine similarity between the document and the
topic string. That measure falls whenever the treatment removes the string
itself, even if the document still covers the topic. A blind judgement of
meaning does not depend on the words the treatment removes.

\subsection{Weak and failed tasks}
\label{sec:negative}

Several tasks and construction choices failed their intended checks. These
failures define where the benchmark is informative and where it is not.

\paragraph{Weak clustering tasks.} Adjusted Rand index is corrected for
chance. Subject clustering separates the model panel, with a median of 0.3439
and a best score of 0.5152. Genre is much weaker, with a best score of 0.0921.
Register peaks at 0.0678, and geographic clustering peaks at 0.0068. The
geographic task does not support model comparison.

\paragraph{A geographic-label defect.} In the first generation, 76.4\% of
documents carrying two geographic tags had tags from different regions. This
was an artefact of independent sampling rather than a property of the text.
The second generation samples compatible tags and checks the result against a
shuffled-label control.

\paragraph{A rewrite increased label leakage.} We tested a more discursive
writing style as a way to reduce lexical prediction. It raised exact copying
of label terms from 29.3\% to 35.5\%. TF-IDF macro-$F_1$ on a 20-subclass pilot task
also rose from 0.639 to 0.724. We rejected the rewrite.

\paragraph{Suppressing topic words reduced coverage.} A prompt instruction
cut topic leakage, but blind judging found lower topic coverage. A gloss-based
variant reduced that loss, although its interval included zero. The released
generation keeps the original prompt because the study does not support a
free improvement in both measures.

\paragraph{Thresholded pair scores were misleading.} The first pair-task
implementation chose a threshold on test labels and reported $F_1$. For genre
form, register, and audience, the median model then produced the all-positive
result: $F_1=0.667$ with accuracy 0.500. The benchmark now uses AUC, which
measures pair ordering without fitting a test-set threshold.

\paragraph{Run order affected the last decimal places.} Unsorted embedding
caches and variable thread counts caused score drift. The evaluator now sorts
texts, fixes thread settings, and records run metadata. The measured drift is
small relative to the reported effects, but older result files do not all
carry the same provenance fields. The release preserves those files and names
the result set used for each analysis.

\section{External validation}
\label{sec:transfer}

\subsection{Transfer of absolute scores}

SHELF is designed to compare models on controlled bibliographic tasks. Its
scores are not intended to predict accuracy on a library's records. We test
that boundary with 21-class subject classification on Project Gutenberg
passages and 4{,}924 English LCSHBench records.

We fit TF-IDF with logistic regression on each corpus and test it on all
three. Macro-$F_1$ is averaged across the 21 classes. This probe has no prior
language training, so its transfer gap measures the corpora rather than model
memory. It uses the full TF-IDF matrix and a grouped 70/30 split on SHELF. The
0.8686 baseline in \Cref{tab:baselines-cls} instead compresses TF-IDF to 256
dimensions and uses the benchmark's stored 60/20/20 split, so the two
in-domain scores are not expected to match.

\begin{table*}[t]
  \centering
  \begin{tabular}{lccc}
    \toprule
    Train $\backslash$ test & SHELF & Gutenberg & LCSHBench \\
    \midrule
    SHELF     & \makecell{\textbf{0.8796}\\[-1pt]\scriptsize[0.88, 0.88]}
              & \makecell{0.3193\\[-1pt]\scriptsize[0.29, 0.34]}
              & \makecell{0.4078\\[-1pt]\scriptsize[0.38, 0.43]} \\
    \addlinespace
    Gutenberg & \makecell{0.2823\\[-1pt]\scriptsize[0.28, 0.29]}
              & \makecell{\textbf{0.5101}\\[-1pt]\scriptsize[0.48, 0.54]}
              & \makecell{0.2135\\[-1pt]\scriptsize[0.19, 0.23]} \\
    \addlinespace
    LCSHBench & \makecell{0.4458\\[-1pt]\scriptsize[0.44, 0.45]}
              & \makecell{0.2800\\[-1pt]\scriptsize[0.25, 0.31]}
              & \makecell{\textbf{0.5559}\\[-1pt]\scriptsize[0.53, 0.58]} \\
    \bottomrule
  \end{tabular}
  \caption{Subject classification, macro-$F_1$. Rows are training corpora and
  columns are test corpora. Intervals resample test documents.}
  \label{tab:transfer}
\end{table*}

The diagonal measures how easy each corpus is within itself. SHELF reaches
0.8796 where Gutenberg reaches 0.5101 and LCSHBench 0.5559, so SHELF is a much
easier classification problem than either natural corpus. That gap is
consistent with the label-term excess measured in \Cref{sec:surface}.

Scores fall in every transfer direction. A probe trained on SHELF scores
0.8796 on SHELF but 0.3193 on Gutenberg, and a probe trained on Gutenberg
scores 0.2823 on SHELF. Corpus size does not explain this: shrinking all three
corpora to the same 3{,}016 documents leaves SHELF at 0.8052 on its own
documents and 0.2341 on Gutenberg. Nor does document length, which differs
sharply across the three corpora (\Cref{tab:lengths}): with every document
cut to its first 200 words, SHELF remains the easiest corpus to
classify within itself, at 0.8521 against 0.4478 for Gutenberg and 0.5522 for
LCSHBench.

Generated text is not the only cause. The two human-written corpora do not
transfer to each other either: a probe trained on Gutenberg scores 0.2135 on
LCSHBench, the lowest cell in \Cref{tab:transfer}. Book passages and catalogue
records differ in form and in cataloguing practice, and that difference
matters even when people wrote both.

Read by column, SHELF is the better off-domain training source for both
natural corpora. On LCSHBench records, a probe trained on SHELF scores 0.4078
where one trained on Gutenberg scores 0.2135; on Gutenberg passages, 0.3193
against 0.2800 from LCSHBench. SHELF and LCSHBench share Library of Congress
vocabulary that literary prose does not. Part of the advantage is size: with
SHELF cut to 3{,}016 documents, its score on Gutenberg falls to 0.2341. The
conclusion is still narrow: a SHELF score describes performance on SHELF, not
the accuracy to expect in production.

\subsection{Transfer of model rankings}
\label{sec:ranking}

Benchmark scores need not transfer for model order to remain useful. We rank
24 distinct embedding models on the subject task in SHELF, Gutenberg, and
LCSHBench. Intervals bootstrap models because the question is whether the
order survives a different model panel.

\begin{table*}[t]
  \centering
  \begin{tabular}{lccc}
    \toprule
    Pair & Spearman & 95\% interval & Kendall \\
    \midrule
    SHELF vs Gutenberg     & \textbf{0.870} & [0.61, 0.98] & 0.761 \\
    SHELF vs LCSHBench     & \textbf{0.765} & [0.43, 0.96] & 0.645 \\
    Gutenberg vs LCSHBench & 0.958 & [0.86, 0.99] & 0.841 \\
    \bottomrule
  \end{tabular}
  \caption{Rank agreement between corpora over 24 distinct embedding models.
  Intervals use 2{,}000 bootstrap resamples of the model set.}
  \label{tab:ranking}
\end{table*}

SHELF gives similar subject-model rankings to both natural corpora, and both
intervals exclude zero. The agreement is lower than the 0.958 correlation
between the natural corpora. The intervals overlap, so the table does not
establish a difference between those correlations.

We then test whether the ranking agreement holds across four subject-task
formulations. Classification agreement was already known at preregistration, on the
smaller model panel available then: $\rho=0.877$ against Gutenberg and
$0.777$ against LCSHBench. The panel has since grown, which is why
\Cref{tab:ranking} reports slightly different values for the same
comparison. Before running retrieval, clustering, or pairs, we froze
the decision rule for the resulting four-formulation claim:
agreement requires $\rho \geq 0.6$ with a lower interval bound above zero,
and at least three of four formulations must pass against both natural
corpora. The dated preregistration record is distributed with the evaluation
repository.\footnote{\url{https://github.com/mjbommar/shelf-benchmark/blob/master/docs/PREREGISTRATION.md}}

\begin{table*}[t]
  \centering
  \begin{tabular}{lcc}
    \toprule
    Formulation & Gutenberg & LCSHBench \\
    \midrule
    Classification & 0.870 [0.61, 0.98] & 0.765 [0.43, 0.96] \\
    Retrieval      & 0.813 [0.59, 0.91] & 0.898 [0.75, 0.96] \\
    Clustering     & 0.916 [0.78, 0.97] & 0.950 [0.85, 0.98] \\
    Pairs          & 0.907 [0.74, 0.96] & 0.942 [0.82, 0.98] \\
    \bottomrule
  \end{tabular}
  \caption{Spearman rank agreement by subject-task formulation, with 95\%
  model-bootstrap intervals. Classification, retrieval, and clustering use
  24 models; pairs use 25.}
  \label{tab:formulations}
\end{table*}

All four formulations pass against both corpora. A preregistered stability
check also supports the clustering row. Across five seeds, the median
pairwise rank correlations are 0.991 on SHELF, 0.932 on Gutenberg, and 0.962
on LCSHBench. One of the ten Gutenberg seed pairs is 0.892, below the 0.90
floor, but the preregistered rule applies the median and all three corpora pass.

The capable-only sensitivity analysis removes both OGBert models and
RoBERTa without sentence-specific training. All eight comparisons still pass. Its weakest cell
is subject classification against LCSHBench at 0.655, with an interval of
$[0.22, 0.93]$. The four formulations reuse labels, documents, and
representations. They show that agreement is not confined to one formulation;
they are not four independent validations.

\section{Scope and limitations}
\label{sec:limits}

SHELF supports comparison among the evaluated models on its released tasks.
Its controlled facets also support diagnosis: a model can be tested on
subject, genre, audience, or register without relying on the correlations
found in catalogue records. The following limits bound that use.

\subsection{Reference points and external validity}

Cataloguers have not annotated a SHELF sample, so the scores have no human
agreement rate or expert ceiling. The benchmark shows how models differ on
its labels, but not whether a remaining error would matter to a cataloguer.
The zero-shot decoder comparison in \Cref{sec:decoders} is a machine reference,
not a substitute for human review.

That decoder comparison is also narrow. It covers subject classification only,
uses four models, and gives the decoders no labelled example while the
encoders are fitted on 37{,}795. It does not test genre form, where encoder
scores are lowest. We therefore do not know whether prompted decoders would
close that gap.

All SHELF documents are model-written, and neural retrievers can prefer
model-written text \citep{dai2024biased}. Using several writing models does not
measure or remove that bias. A matched study of human and model-written text
is still needed.

The external comparison covers subject labels only because Gutenberg and
LCSHBench do not provide the other SHELF facets at comparable coverage.
Agreement on subject classification, retrieval, clustering, and pairs does
not validate the genre, audience, register, geography, or instruction tasks.
The transfer matrix also uses one lexical probe. Its scores do not carry
between the three corpora, but another classifier may lose a different amount.

\subsection{Corpus and split provenance}

The main results pool 42{,}532 original records with document-level splits and
20{,}367 later records with specification IDs. The grouped-split guarantee
therefore applies only to the later component. We cannot test whether versions
of one original brief appear on both sides of a split. A complete core-only
sweep under grouped splits would show how much this uncertainty affects the
headline results.

The pooled corpus is also not balanced across writing models: its largest
model supplies 47.7\% of the documents. Comparisons among writing models must
use the balanced factorial component described in \Cref{sec:corpus}. The
released pair tasks come from the original corpus component, as stated in
\Cref{tab:pairs}.

\subsection{Models, tasks, and implementation}

The evaluated encoders range from 22M to 596M parameters. This includes recent
small and mid-size models, but not the billion-parameter systems at the top of
general embedding leaderboards \citep{elassadi2026dilemma}. A larger panel may
produce a different order.

The evaluator omits the query and document prefixes recommended by individual
model cards. One input format makes the comparison consistent, but it may
favour models that need less prompting. These rankings apply to the unprompted
protocol, not to each model's best configuration.

Several tasks do not separate the current models. Geographic clustering and
the register and audience pair tasks remain at or near chance. They work as
negative controls, but not as useful leaderboards.

Document length differs across models and corpora. Many encoders cut text at
512 tokens, while SHELF documents often exceed that limit. Running
GTE-ModernBERT at 2{,}048 and 8{,}192 tokens changed the mean across 32
task--corpus cells by $-0.0006$. Longer context did not help that model, but
this experiment does not test the step from 512 tokens or any other encoder.

Finally, the timing results in \Cref{tab:timing} come from one run and have no
repeated-trial intervals. TF-IDF uses CPU, encoders use GPU batches, local
decoders process one document at a time, and the hosted model uses concurrent
requests on undisclosed hardware. TF-IDF reads whole documents while neural
inputs have token limits. These are observed processing rates for one harness
and machine, not hardware-neutral measures of latency, cost, or energy.

\section{Conclusion}
\label{sec:conclusion}

SHELF tests bibliographic work that general leaderboards do not isolate. This
release contains 62{,}899 documents and tasks for classification, clustering,
retrieval, pair classification, and instruction retrieval. The results show
why those tasks should remain separate. Current methods classify broad
subjects well, yet struggle with fine genre forms, register, audience, and
several clustering tasks. Sparse methods remain strong classifiers, while
dense encoders lead retrieval. In the timing experiment, TF-IDF has a
near-best subject score and the highest measured processing rate.

The controlled data make these differences easier to diagnose. SHELF can vary
subject, genre, audience, register, and writing model independently, and its
factorial subset includes combinations unlikely to occur in most natural
collections. But control does not make synthetic scores equivalent to
production accuracy. Label terms make SHELF somewhat easier, some tasks remain
near chance, and the corpus has no human catalogue judgement as a reference.

The comparisons with Project Gutenberg and LCSHBench sharpen that boundary.
Absolute scores change sharply across corpora, while subject-model rankings
are more stable. SHELF can therefore help choose and compare methods on its
released tasks, but it cannot predict accuracy on a library's catalogue. A
future study with cataloguer annotations and a broader encoder panel would
test both limits directly.

SHELF is a system as well as a fixed release. It can generate new documents
after a future model's training cutoff while preserving the labels and task
design. We release the corpus \citep{shelf2026}, source code, per-task results,
and scripts under permissive licences on Hugging Face and at
\url{https://github.com/mjbommar/shelf-benchmark}.

\section*{Disclosures and limitations}
\addcontentsline{toc}{section}{Disclosures and limitations}

\begin{description}[leftmargin=!, labelwidth=0.23\linewidth,
                    font=\normalfont\bfseries\color{primary},
                    itemsep=0.3em, style=nextline]
  \item[Data availability] The dataset is available on
    \href{https://huggingface.co/datasets/mjbommar/SHELF}{Hugging Face}.
    Evaluation code and stored per-task results are available on
    \href{https://github.com/mjbommar/shelf-benchmark}{GitHub}.
  \item[Funding] \PaperFunding.
  \item[Competing interests] \PaperCompeting.

    \item[AI assistance] \PaperAIStatement

  \item[Limitations] The limits on interpretation, including the absence of a
    human reference and a frontier-scale encoder, are stated in
    \Cref{sec:limits}.
\end{description}

\PrintBibliography

\end{document}